\documentclass[letterpaper]{article}
\usepackage{aaai}
\usepackage{times}
\usepackage{helvet}
\usepackage{courier}
\usepackage{threeparttable}
\usepackage{makecell}
\usepackage{multirow}
\usepackage{amsfonts}
\usepackage{graphicx}
\usepackage{color}
\usepackage{amsthm}
\usepackage{amssymb}
\usepackage{amsmath}
\usepackage{arydshln}
\usepackage{amsmath}
\usepackage{algorithm}
\usepackage[noend]{algpseudocode}
\usepackage{algorithmicx}
\usepackage{microtype}
\usepackage{stfloats}
\usepackage{multirow}
\usepackage{makecell}
\usepackage{MnSymbol}
\usepackage{mathtools}
\usepackage{url}

\usepackage{amssymb}
\usepackage{pifont}

\newcommand{\ml}{Spectro-Temporal Transformer}
\newcommand{\ms}{STT}
\newcommand{\dl}{WildMix}

\DeclareMathOperator*{\argmin}{argmin}

\begin{document}
%
\title{WildMix Dataset and Spectro-Temporal Transformer Model \\ for Monoaural Audio Source Separation}
\author{Amir Zadeh$^\dagger$, Tianjun Ma$^\dagger$, Soujanya Poria$^*$, Louis-Philippe Morency$^\dagger$\\
$^\dagger$ Language Technologies Institute, School of Computer Science, Carnegie Mellon University\\ $^*$ Singapore University of Technology and Design  \\
\texttt{\{abagherz,tianjunm,morency\}@cs.cmu.edu},\texttt{soujanya\_poria@sutd.edu.sg}}
\maketitle

\begin{abstract}
\begin{quote}
Monoaural audio source separation is a challenging research area in machine learning. In this area, a mixture containing multiple audio sources is given, and a model is expected to disentangle the mixture into isolated atomic sources. In this paper, we first introduce a challenging new dataset for monoaural source separation called \dl. WildMix is designed with the goal of extending the boundaries of source separation beyond what previous datasets in this area would allow. It contains diverse in-the-wild recordings from $25$ different sound classes, combined with each other using arbitrary composition policies. Source separation often requires modeling long-range dependencies in both temporal and spectral domains. To this end, we introduce a novel trasnformer-based model called Spectro-Temporal Transformer (STT). STT utilizes a specialized encoder, called Spectro-Temporal Encoder (STE). STE highlights temporal and spectral components of sources within a mixture, using a self-attention mechanism. It subsequently disentangles them in a hierarchical manner. In our experiments, STT swiftly outperforms various previous baselines for monoaural source separation on the challenging \dl \ dataset. 
\end{quote}
\end{abstract}
\section{Introduction}

Disentangling auditory sources is both a vital capability for future AI (artificial intelligence) systems, and a fundamental challenge in the field of machine learning. In the real world, AI systems need to cope up with sound complexities happening around them. For example, a dialogue system should not fail just because the agent's microphone can pick surrounding sounds in the vicinity of the conversation. Monoaural source separation, where all overlapping audio source share the same channel, is arguably the most challenging scenario for audio source separation. In this scenario, a machine learning model is given a mono-channel sound mixture, and is expected to generate the atomic constituent sources. Despite recent advances in deep learning, monoaural source separation remains largely understudied due to both lack of datasets with large variety within mixtures, and lack of efficient models to capture very long-range dependencies often required for extracting sources.
\begin{figure}[t!]
\begin{center}
\includegraphics[width=.8\linewidth]{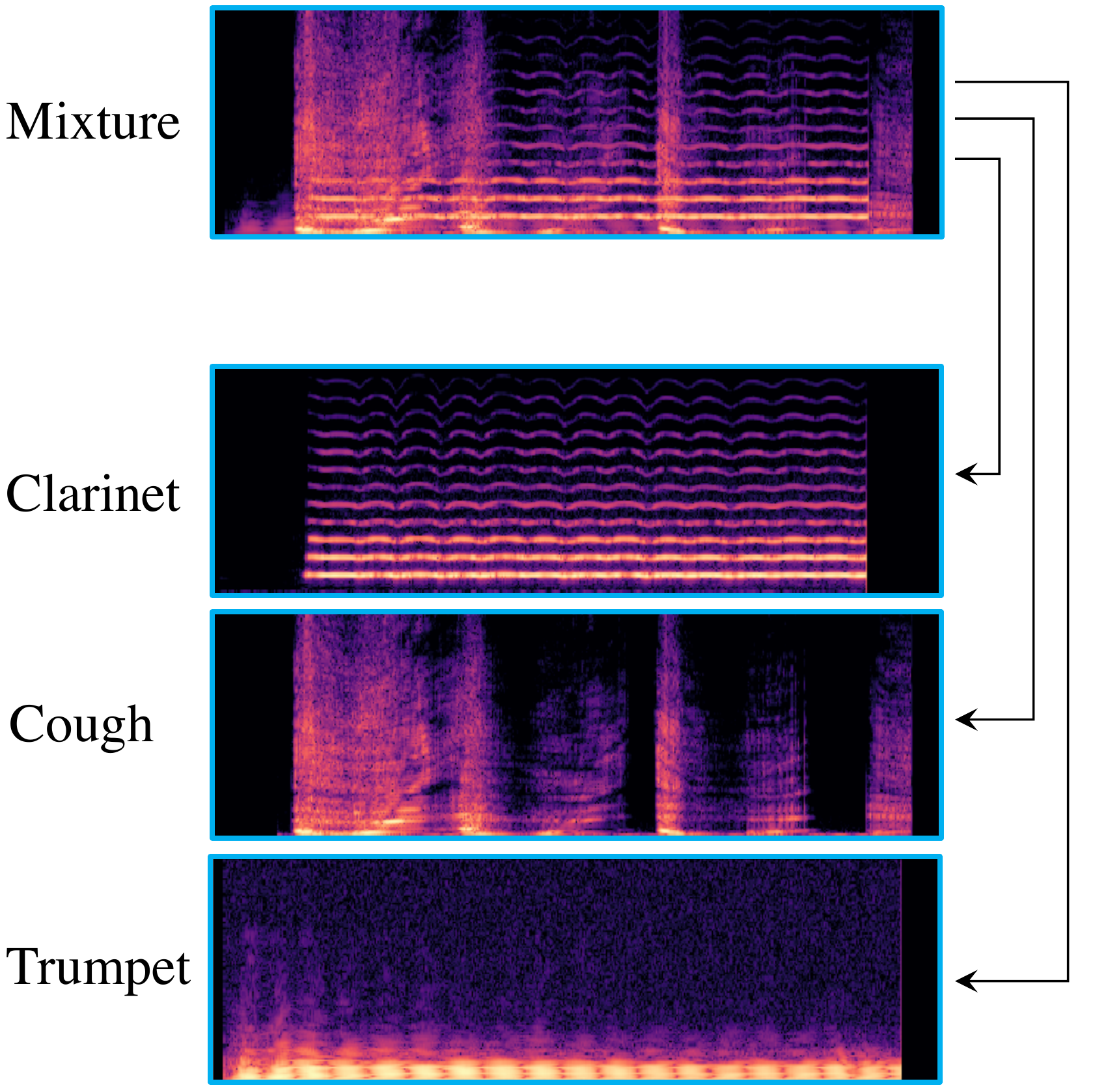}
\end{center}
\caption{Example of a sound mixture and its constituent sources. Each source has arbitrary short and long range dynamics, scattered across temporal and spectral domains. Example is taken from \dl \ dataset. }
\label{fig:teaser}
\end{figure}

Diversity is a crucial factor for a monoaural sound separation dataset. In many real-world scenarios, arbitrary sound sources can overlap together, forming complex and challenging mixtures. These complex mixtures go beyond classical Cocktail Party (i.e. human speech overlap, or background/foreground separation) or music instrument separation often studied in previous works (further discussed in Related Works section). A dataset that allows for in-depth studies in monoaural source separation should reflect this  diversity found in natural scenarios. To this end, we introduce \dl \ dataset: a dataset with $25$ different sound classes. The design of \dl \ allows for studying overlapping sources in challenging setups, further discussed in continuation of this paper. 

A sound mixture is inherently a form of spatio-temporal data (with the spatial domain being the spectral coefficients). For each source within the mixture, there are spatio-temporal relations that are unique to the source class (e.g. dog barking, keys jingling, or human voice). Figure \ref{fig:teaser} shows an example of a mixture and three underlying sources. Each source has arbitrarily scattered (yet related) coefficients in both temporal and spectral domains of the mixture. To recreate sources within a mixture, these scattered coefficients need to be highlighted and extracted from the mixture; a task which is non-trivial and requires a model capable of finding arbitrary dependencies. Transformer models \cite{vaswani2017attention} are a unique class of neural models for extracting arbitrary compositions (short and long-range) within their input space \footnote{In some cases with sequence length as long as $512$~\cite{devlin2018bert}}. Their superior performance over RNN-based models is credited to a self-attention mechanism. Self-attention has a full input-size receptive field (thus allowing for looking at the entire input in one pass) as opposed to relying on a recurrent architecture (which is notoriously hard for modeling long range sequences due to numerical or optimization issues~\cite{pascanu2013difficulty}). This is a particularly appealing feature for highlighting sources within a mixture, and subsequently disentangling them. \ml\footnote{Name inspired from Spectro-Temporal Receptive Field (STRF) in human brain, where auditory neurons are sensitive to certain frequency and time patterns\cite{richard1995rapid}.} \ builds upon this self-attention mechanism and uses a proposed Spectro-Temporal Encoder (STE) to highlight relevant rows (spectral domain) and columns (temporal domains) within an input spectrogram. Our experiments study the performance of STT and several baselines over \dl.

\section{Related Works}
\newcolumntype{K}[1]{>{\arraybackslash}p{#1}}
\begin{center}
\renewcommand{\arraystretch}{1}
\begin{table}[t!]
\fontsize{9}{10}\selectfont
\centering
\begin{tabular}{K{2.5cm}|c|c|K{1.2cm}}
\hline
Dataset   & Year     & \#Class          & Domain         \\ \hline
TIMIT   & 1986    &     1      &   CP   \\ \hline
WSJ0   & 1993    &      1     &   CP   \\ \hline
RWC   & 2003    &      2-4     &   Music   \\ \hline
MASS & 2004 &    2-4  & Music               \\ \hline
VCTK     & 2009 &    1  & CP  \\ \hline
DREANSS     & 2013 &   2   & Music  \\ \hline
MIR-1k & 2013     &  2-4 &   Music    \\ \hline
TSP & 2014     & 1 &   CP    \\ \hline
DT & 2015     &    1    &    CP      \\ \hline
iKala & 2015     &     2-4   &    Music      \\ \hline
DSD100 & 2016     &   5  &     Music  \\ \hline
VoiceBank & 2016     &  1   &     CP  \\ \hline
JADE & 2017     &   1   &  CP \\ \hline
MUSDB18 & 2018     &   5   &  Music \\ \hline
AVSpeech & 2018     &   1   &  CP \\ \hline
MUSDB-HQ & 2019     &  5 &     Music    \\ \hline
\hline
\dl \ (ours)& 2020     & 25  & Diverse \\ \hline
\end{tabular}
\caption{\label{table:otherdata}A list of well-known source separation datasets in the literature. CP refers to Cocktail Party scenario.}
\end{table}
\end{center}
The related works to the material in this paper are split in the following two areas:

\subsection{Monoaural Source Separation Datasets:} Datasets that contain both mixtures and their corresponding source tracks have received a particular attention in monoaural source separation. There are several well-known datasets in this area, some of which have influenced \dl \ dataset design. Table \ref{table:otherdata} lists these datasets, which mainly fall into the following two categories: 

\noindent \textit{Cocktail Party} scenario, where mixtures constitute of overlapping human voices. The most well-known datasets in this area are: TIMIT \cite{zue1990speech}, WSJ0 \cite{WSJ0}, VCTK \cite{weinberger2009towards}, TSP \cite{huang2014deep}, DT \cite{simpson2015deep}, VoiceBank \cite{valentini2016investigating}, JADE CPPdata \cite{miettinen2017blind} and  AVSpeech \cite{ephrat2018looking}. In majority of these datasets, there is only one class of sound, which is human voice. Most of the above datasets have only 2 sources within each mixture (2 voices in each mixture). 

\noindent \textit{Music} separation, where the task is to separate the music instruments within a given mixture. The most well-known datasets in this area are
MIR-1k \cite{hsu2012tandem}, RWC dataset \cite{goto2002rwc}, MASS \cite{MTGMASSdb}, DREANSS \cite{marxer2013study}, MUSDB18\cite{musdb18}, MUSDB-HQ \cite{musdb18-hq}, DSD100 \cite{SiSEC16}, iKala\cite{chan2015vocal}. Majority of these datasets focus on 4 instrument classes of Vocal, Drum, Bass, and Others (all the other accompaniments are considered as a separate class). Some of the music separation datasets, such as MIR-1K, MUSDB18, allow for voice (singing) vs background separation.

Majority of the above datasets use audio overlaying to combine individual sources into a mixture (e.g. violin, bass and vocal tracks independently recorded and ultimately compiled into a song). While previously proposed datasets have mainly focused on the two scenarios of Cocktail Party and music separation, with limited variations in sound classes, \dl \ dataset allows for research in higher number of intermixed classes and more challenging setups, further discussed in the continuation of this paper. 

\subsection{Monoaural Source Separation Models}
Audio source separation has been among the ambitious goals of AI for a few decades \cite{lee1997blind}. Aside non-paramteric models which rely mostly on feature engineering \cite{wood2017blind}, parametric models have been widely used for monoaural source separation. Following a supervised setup, a mixture is given to a model and the atomic separated sources are expected as the output (with supervision of number of sources in the mixture - but no supervision the source classes). With the advent of deep neural networks, neural approaches \cite{wang2008time,grais2014deep,weninger2014single,huang2015joint} have become popular due to their superior performance over traditional non-parametric or non-neural approaches. Specifically, recurrent neural networks (with a particular focus on LSTMs) have provided a stepping stone for several source separation models courtesy of their sequence modeling capabilities \cite{chen2017long,sun2019monaural}. However, RNNs are in many cases unsuccessful in modeling long sequences, as they are prone to numerical and optimization problems \cite{pascanu2013difficulty}. This can pose challenges to audio source separation, which requires modeling arbitrary and often long-range dependencies efficiently (e.g. Figure \ref{fig:teaser}).

\section{\dl \ Dataset}
\label{sec:daq}
In this section, we introduce the \dl \ dataset\footnote{Dataset will be publicly available for download after May 30th, 2020.}. Mixtures in \dl \ dataset contain a variety of naturally occurring sound classes. The sounds are combined using different strategies, making \dl \ both a challenging dataset for future research, and a unique resource for detailed studies in source separation. We first start by outlining the data acquisition process, followed by mixture creation procedure. \begin{center}
\renewcommand{\arraystretch}{1.5}
\begin{table*}[t!]
\fontsize{9}{10}\selectfont
\centering
\begin{tabular}{K{3.1cm}|K{3.1cm}|K{3.1cm}|K{3.1cm}|K{3.1 cm}}
\hline
1. Speech   & 2. Cowbell     & 3. Meow          & 4. Violin         & 5. Typing \\ \hline
6. Guitar   & 7. Laughter    & 8. Oboe          & 9. Keys Jingle    & 10 .Cough             \\ \hline
11. Applause & 12. Finger Snap & 13. Snare Drum    & 14. Shatter        & 15. Saxophone         \\ \hline
16. Bark     & 17. Flute       & 18. Paper Tearing & 19. Pencil Writing & 20. Knock             \\ \hline
21. Clarinet & 22. Gunshot     & 23. Trumpet       & 24. Tambourine     & 25. Electric Piano    \\ \hline
\end{tabular}
\caption{\label{table:sources}Sound classes within \dl \ dataset. For further description, please refer to \dl \ Dataset Section. }
\end{table*}
\end{center}
\subsection{Data Acquisition}

Our data acquisition can be summarized in two stages: 1) \textbf{Class Selection:} selection of a diverse set of naturally occurring sound classes, 2) \textbf{Sound Verification and Diversity:} creating diversity within each sound class.  \newline

\noindent \textbf{Class Selection:} Table \ref{table:sources} shows all the sound classes chosen for \dl \ dataset. There are a total of $25$ diverse sound classes including human sounds, animal sounds, music instruments, and object sounds. The diversity among these sound classes creates a challenging environment for the source separation models. Note that from hereon, we use the term ``class'' to refer to entries in Table \ref{table:sources}.\newline

\noindent \textbf{Sound Verification and Diversity:} For each of the classes in Table \ref{table:sources}, our goal is to acquire a set of diverse atomic recordings which represent the class (e.g. diverse human voices). For each class, we query the Freesound \footnote{https://freesound.org - a free website for community-recorded royalty-free music. Gathered audios follow creative-commons license.} website (with the exception of human speech) for user-recorded in-the-wild audio segments. We manually verify the sound class, check the sound quality and verify the audio segment being atomic (no other sound than the desired class). We also manually check for intraclass diversity (to ensure no two sounds are identical or very similar to each other). For the human speech, we choose recordings from the CMU-MOSI dataset~\cite{zadeh2016mosi}. CMU-MOSI is a gender-balanced monologue dataset containing voices of $89$ distinct speakers. We gather $60$ audio segment for each class in Table \ref{table:sources}, with lengths of $0.25$ to $1$ seconds. The $60$ recordings in each class subsequently fall into train ($40$), validation ($10$) and test ($10$). No audio segment is shared between these folds. 

In summary, our data acquisition allows for creating a dataset that is diverse, not just across classes but also within classes.

\subsection{Mixture Creation}
Mixture creation is at the core of the \dl \ dataset. Our goal is to create a challenging mixture that pushes the boundaries of source separation and sparks further research in this area. To do so, we adopt the following policies: \textit{Arbitrary Composition} - In a sound mixture, there should be no temporal dependency between the sources, e.g. no persistent pattern of one class starting before or after another class. Furthermore, there should be no co-occurring bias between the sources (e.g. violin always accompanied with only music instruments). In an unbiased scenario, the classes are considered to be i.i.d. \textit{Arbitrary Volume} - The process of convolving sounds into a mixture should follow a random volume procedure. In real scenarios, sources may come at different volumes. For example, coughing sound may have lower or higher volume than a laughter. Therefore, the mixture should reflect this diversity and consider the volume to be a random parameter. \textit{Under-determinism} - The sources within a mixture should come from different microphones with different intrinsic parameters. If all the sounds are recorded with similar microphones, then models may not generalize in real world. Data acquisition of user-recorded in-the-wild sources allows for this diversity to be naturally captured since different devices are used to record sounds. The creation of the \dl \ dataset closely follows these policies. The audio files are mixed together using ffmpeg, which mimics the natural process of audio sources overlaying. What follows is the formalization of the mixture creation, accompanied by terminologies and definitions. 

Let ${A}=\{A_i;A_i=\{a_{<i,j>}\}_{i=1}^{N_a}\}_{j=1}^{N_A}$ be the set of audio segments of different classes in Table \ref{table:sources}, with $N_A=25$ being the total number of classes and ${N_a}=60$ the number of audio segments for each class ($60$). Acquired audio segments are all high quality and sampled at $44.1$ KHz - aligned with the most recent trend of keeping mixtures high quality~\cite{musdb18-hq}. The recordings in each class are split in three sets of train ($40$ sources), validation ($10$ sources) and test ($10$ sources). These sets are mutually exclusive (i.e. no source is shared between these sets). Based on how the sound classes are chosen for being mixed, \dl \ is split into 3 partitions: a) \textbf{Interclass:} where overlapping sources are chosen strictly from different classes for each mixture, b) \textbf{Intraclass:} where, for each mixture, overlapping sources are strictly from the same class, c) \textbf{Hybrid:} where overlapping sources can be from the same or different classes. Let $U=\{Interclass, Intraclass, Hybrid\}$ denote the set of partitions from hereon. Each of the partitions in turn consists of 3 tasks based on number of underlying audio segments present in the mixture. Let $S=\{2,3,5\}$ denote the set of tasks. 

Each subdataset (combination of partition and task) is identified with a tuple $<u \in U, s \in S>$. Therefore, there are $9$ subdatasets within the full \dl \ dataset. For each subdataset there are $3$ folds, $F=\{tr,vl,te\}$ - train, validation and test. The data within the subdataset $<u,s>$ is denoted as ${X}_{<u, s>}=\{\big[X_{<u,s,f>}[i]\big]_{i=1}^{D_{s,f}}\}_{f\in F}$. Note that $D_{s,f}=(10^4 \times s)$ for $f=tr$ (train set) and $D_{s,f}=(10^3 \times s)$ for $f \in \{te,vl\}$ (validation and test). Depending on the subdataset $<u \in U, s \in S>$, $s$ audio segments are chosen based on $u$ and randomly assigned a volume and a start time. Algorithm \ref{alg:mixture} in supplementary formalizes the process of creating the mixture. 

All the mixtures ${X}_{<\cdot>}[\cdot]$ in \dl \ dataset have the length of $2$ seconds \footnote{Longer sequences can be tiled and separated every 2 seconds.}. Ultimately, the goal of separation is to extract $\bar{X}_{<u,1:s>}[\cdot]$, which are the separated $1:s$ sources, given a mixture ${X}_{<u,s>}[\cdot]$. From hereon, bar above $X$ is used to denote the individual sources in the mixture. Note we use the term ``source'' (and not audio segment) for separation results. The source $\bar{X}_{<\cdot>}$ has the same time duration as the mixture $X_{<\cdot>}$, but underneath there is an atomic audio segment $a_{<\cdot>}$ padded (if needed) at the beginning/end by silence and randomly volumized. While sources and mixtures are based on PCM (Pulse Code Modulation) values, the spectrogram representation is subsequently obtained using STFT(Short-Time Fourier Transform) for experiments. Similar to images, spectrograms have a width $W$ and a height $H$ (the temporal and spectral space respectively). With a small redefinition, after mixture is created, we use ${X}_{<\cdot>}$ and $\bar{X}_{<\cdot>}$ to refer to spectrograms and not the PCMs (since \ms \ and baselines all use spectrograms as input). Hence ${X}_{<\cdot>}[\cdot], \bar{X}_{<\cdot>}[\cdot] \in \mathbb{R} ^{W \times H}$. In this paper, we choose a Hann window of size $256$, and hop length of $192$, which leads to 
$W=460$ and $H=258$ ($129$ real and $129$ imaginary spectral coefficients concatenated). 

\section{Spectro-Temporal Transformer}
\begin{figure}[t!]
\begin{center}
\includegraphics[width=.75\linewidth]{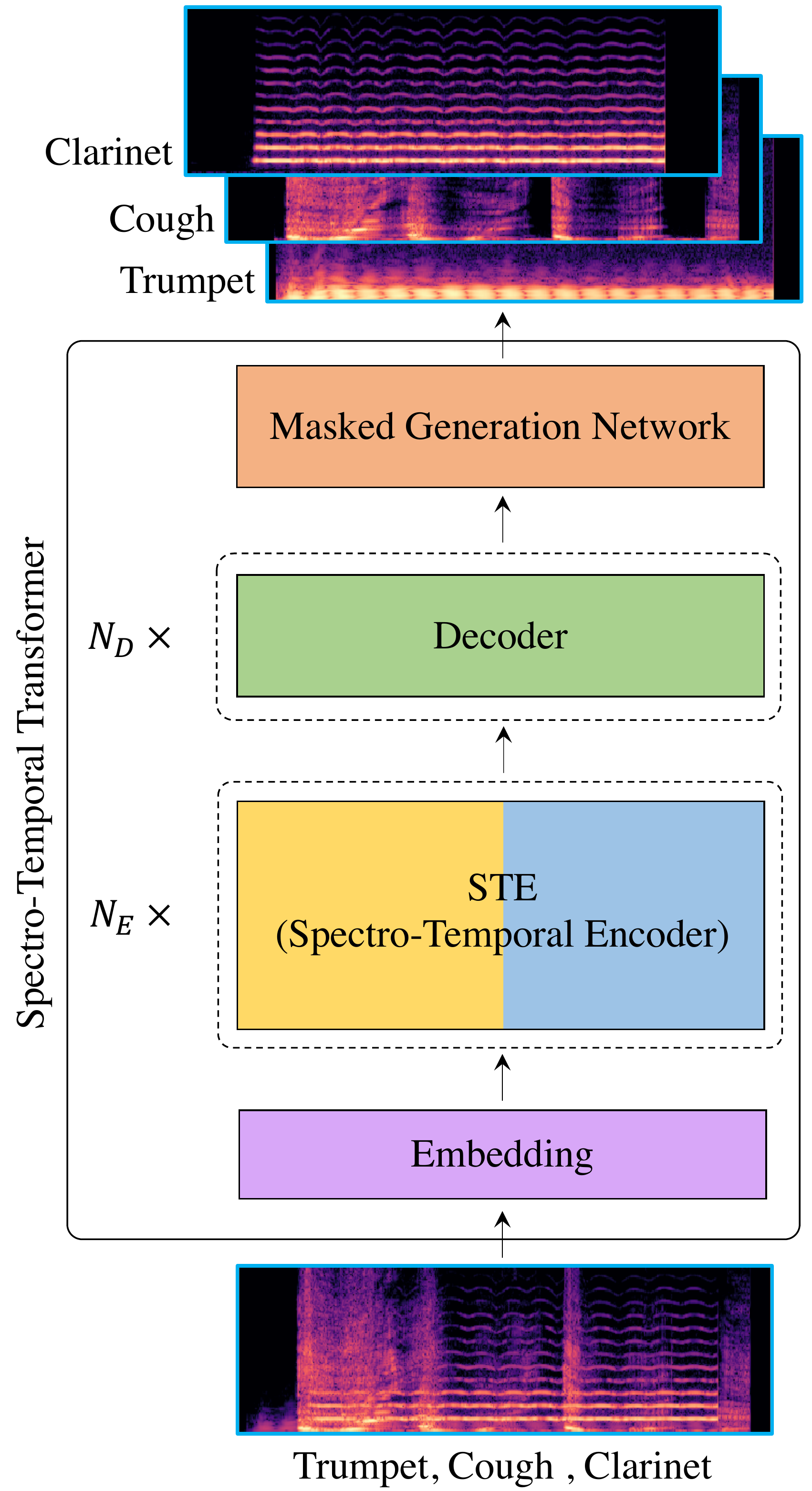}
\end{center}
\caption{Overview of the \ms \ (\ml) model. $N_E$ denotes the number of STE in the encoder stack and $N_D$ denotes the number of decoders in the decoder stack.}
\label{fig:sttfull}
\end{figure}
In this section we outline the proposed \ms \ (\ml). \ms \ is a transformer-based model tailored for hierarchically separating auditory sources in a mixture. It extracts relations (short and long) within a mixture from both temporal and spectral domains, using temporal and spectral self-attention mechanisms. Figure \ref{fig:sttfull} shows an overview of the operations within the \ms \ model. At the input of \ms, there is an Input Embedding Layer. Afterwards, \ms \ follows an encoder-decoder architecture with the following main components: 1) \textbf{STE (Spectro-Temporal Encoder)}, a specialized temporal and spectral encoder designed to disentangle the sources within the mixture. 2) \textbf{Decoders} which receives the output of the STE stack and proceed to build the sources, followed by a 3) \textbf{Masked Generation Network (MGN)} which generates the final sources. We discuss each of these components briefly in the continuation of this section. Algorithm \ref{alg:model} in supplementary, outlines the exact operations of the \ms \ and its underlying components. For exact Pytorch implementation, \ms \ code will be released publicly after Feb 7th, 2020. 

\subsection{Input Embedding Layer}

For $i$th datapoint\footnote{For simplicity of notation, we discuss the operation for an individual datapoint. In practice all the operations are done in batch form.}, the input to the \ms \ is a mixture spectrogram $X_{<u \ \in \ U,s\ \in \  S,\cdot>}[i] \in \mathbb{R}^{W \times H}$ and the output is a separated mixture $\{\bar{X}_{<u,1:s,\cdot>}\}[i] \in \mathbb{R}^{W \times H}$ containing only individual sources. Given the input spectrogram $X_{<u,s,\cdot>}[i]$, we first use an embedding network with positional information \cite{vaswani2017attention} in both temporal and spectral domains. This network embeds each column (the spectral dimension) of the spectrogram into a canonical shape $H_e$ for the subsequent encoder and decoder stacks. We regard the output of this embedding network as $X^{Em}_{<u,s,\cdot>}[i] \in \mathbb{R}^{W \times H_e}$.

\begin{figure}[t!]
\begin{center}
\includegraphics[width=0.75\linewidth]{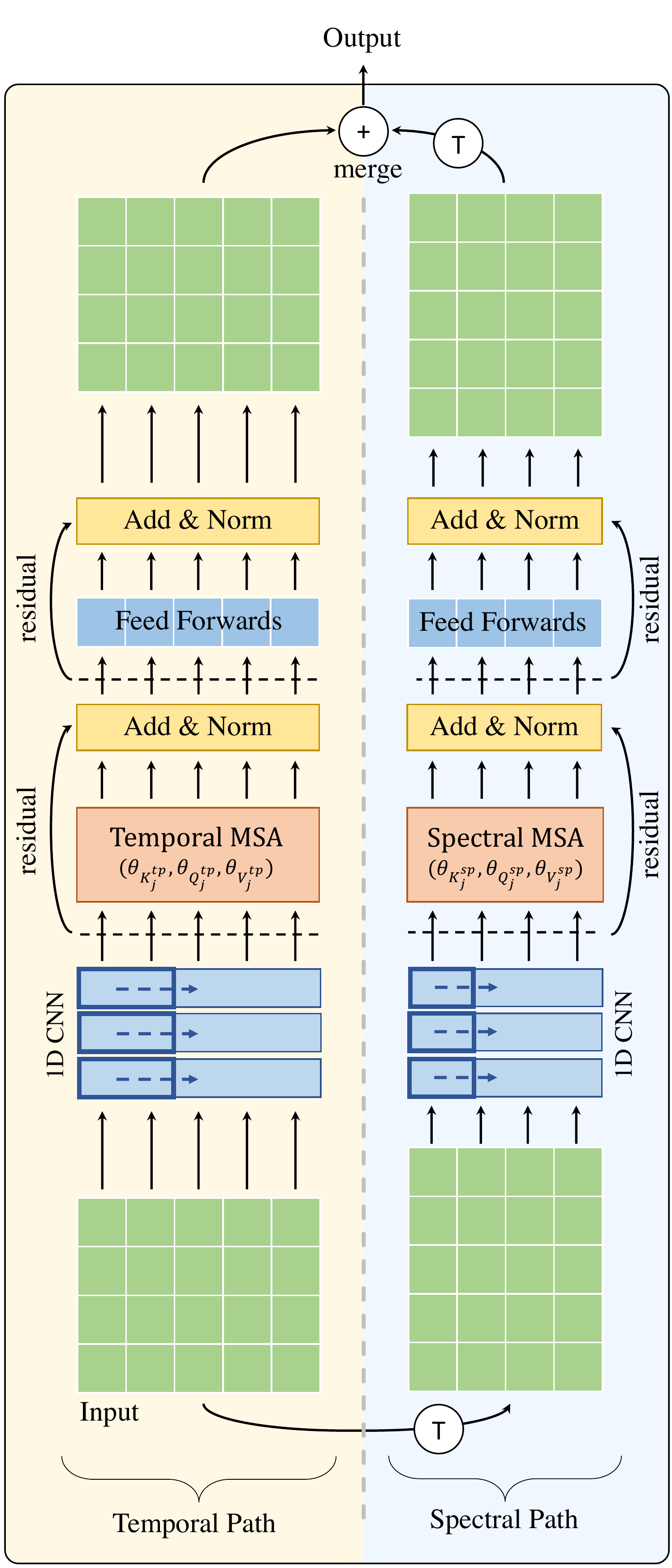}
\end{center}
\caption{Best viewed zoomed in and in color. Overview of the Spectro-Temporal Encoder (STE). There are two attention components: Temporal MSA (Temporal Multi-head Self Attention) which is an attention over the temporal dimensions of the input and Spectral MSA (Spectral Multi-head Self Attention) which is an attention over the spectral dimensions.}
\label{fig:encoder}
\end{figure}

\subsection{Spectro-Temporal Encoder (STE)} The stack of STE ($N_E$ total STE in the stack) receives the output of the embedding layer $X^{Em}_{<u,s,\cdot>}[i]$. Figure \ref{fig:encoder} summarizes the operations done within the $j$th STE. Let $X^{E(j)}_{<u,s,\cdot>}[i]$ be the input to the $j$th STE (with $X^{E(1)}_{<u,s,\cdot>}=X^{Em}_{<u,s,\cdot>}[i]$). Inside STE, there exists two paths: a temporal path and a spectral path. The temporal path disentangles sources within the mixture using operations on the temporal space. Similarly, spectral path disentangles sources within the mixture using operations on the spectral space. The input to the $j$th STE is always within the temporal domain. We transpose this input for the spectral path. The first operation within each path is a set of 1D deep convolutions which disentangle the sources with respect to their path domain (temporal or spectral). The architecture of this 1D CNN is a hyperparameter of the \ms \ model. The output of the CNN relies on the same space as STE input by using valid convolutions (no pooling layers). We denote the output of this CNN operation as $X^{CNN-tp(j)}_{<u,s,\cdot>}  \in \mathbb{R}^{W \times H_e}$ for temporal, and $X^{CNN-sp(j)}_{<u,s,\cdot>}  \in \mathbb{R}^{H_e \times W}$ for spectral path. There are two attention components within the STE: one Temporal MSA (Multi-head Self Attention, \cite{vaswani2017attention}) within temporal path, and one Spectral MSA within the spectral path. These attentions are essentially the components that highlight the source dynamics across the temporal and spectral domains of the mixture. The Temporal MSA is controlled by key $K^{tp}_{j}$, query $Q^{tp}_{j}$ and value $V^{tp}_{j}$. Similarly the Spectral MSA is controlled by key $K^{sp}_{j}$, query $Q^{sp}_{j}$ and value $V^{sp}_{j}$. The output of MSA in each path is added with its input by a residual connection and followed by a normalization layer \cite{vaswani2017attention}. Subsequently, the output of normalization layer in each path goes through a set of feedforward networks (one per each column of the input). It is afterwards residually added with the output of the feedforward networks and normalized again. Finally, the temporal and spectral paths merge by transposing the output of spectral path and adding it with the output of the temporal path. 

\subsection{Decoder and Masked Generation Network}
The output of the final STE layer is passed to the decoder stack to recreate the the individual sources. We use a similar decoder architecture as the original transformer model~\cite{vaswani2017attention}. There are a total of $N_D$ decoders in the stack, all of which have outputs that lies in $\mathbb{R}^{W \times H_e}$. The output of the final decoder is subsequently used as input to the Masked Generated Network (MGN), to generate the final sourcess. The input of MGN first goes through a feedforward network to map the decoder output from $\mathbb{R}^{W \times H_e}$ to $\mathbb{R}^{W \times H \times s}$. Subsequently, the output of this feedfroward goes through another feedforward network to get a ReLU activated (non-negative) output mask, also in $\mathbb{R}^{W \times H \times s}$. The mask is then elementwise multiplied with the output of the first feedforward to generate the final output. In practice, we found that this masking is important for generation performance (see Results and Discussion section). The architecture of the two feed forward networks in MGN are hyperparameters of the model.

\section{Experiments}
In this section we first describe the baselines used as points of comparison to \ms. We then proceed with outlining the experimental methodology including loss function and hyperparameter choices. 

\subsection{Baselines}
The following baselines are compared with each other for all the subdatasets $<u \in U, s \in S>$ of the \dl \ dataset. We implement each baseline based on published code by the original authors (or we implement the to the best of our knowledge if code is not published). All the baseline models in their original format (as well as \ms) expect supervision of number of sources (but no supervision of the source classes). This supervision in turn is used to change their output layers to generate $s \in S$ sources. In this paper, we focus on generic source separation on all the \dl \ subdatasets and not a particular scenario (e.g. Cocktail Party). 

 \textit{DNN} is a baseline that uses a fully connected deep neural network for separating the sources within a mixture \cite{grais2014deep}. 

 \textit{DRNN/SRNN} are baselines that uses two types of RNN to simultaneously model all sources \cite{huang2015joint}. 

\textit{SSP-LSTM} is a deep LSTM model designed for speech source separation. Aside training the orignal SSP-LSTM, we try the bidirectional variant of this model as well. Unlike other baselines, SSP-LSTM does not have any particular component at the final source generation stage. \textit{MGN-LSTM} is a deep LSTM baseline model designed in this paper. It uses Generation Residual at the outuput of the SSP-LSTM, to strengthen the final generation component of the SSP-LSTM. 

\textit{CSA-LSTM:} is a Complex Signal Approximation baseline that focuses on careful generation of the complex domain of the spectrogram during training \cite{sun2019monaural}.

\textit{L2L:} is a strong source separation baseline which uses deep dilated convolutions and a bidirectional LSTM \cite{ephrat2018looking}. The original paper contains a competative audio-only implementation, which is used here. 

\textit{OTF} is the implementation of the original transformer model \cite{vaswani2017attention}. This model does not have the STE, but rather the original proposed encoder. To generate the output, we use a Generation Residual layer at the end of decoder stack. 
\begin{table*}[t!]
\begin{center}
\fontsize{9}{10}\selectfont
\renewcommand{\arraystretch}{1.2}

\begin{tabular}{K{5.25cm}|K{0.9cm}|K{0.9cm}|K{0.9cm}|K{0.9cm}|K{0.9cm}|K{0.9cm}|K{0.9cm}|K{0.9cm}|K{0.9cm}}
\hline
\multirow{2}{*}{\textit{Baseline}} & \multicolumn{3}{c|}{$u=$\textit{Interclass}}     &  \multicolumn{3}{c}{$u=$\textit{Intraclass}}  &  \multicolumn{3}{|c}{$u=$\textit{Hybrid}} \\ \cline{2-10} 
 & $s=2$ & $s=3$ & $s=5$ & $s=2$ & $s=3$ & $s=5$ & $s=2$ & $s=3$ & $s=5$ \\ \hline
Mixture Projection (worst case) & 15.000 & 16.500 & 16.542 & 15.330 & 15.730 & 14.005 & 15.631 & 17.150 & 16.042 \\ \hline
DNN \cite{grais2014deep} & 15.060 & 14.802 & 15.380 & 15.418 & 15.337 & 15.423 & 15.680 & 15.364 & 15.140  \\ \hline
DRNN \cite{huang2015joint} & 15.031 & 14.802 & 15.381 & 15.510 & 15.377 & 14.546 & 15.668 & 15.359 & 15.125 \\ \hline
SRNN \cite{huang2015joint} & 12.733 & 12.533 & 13.651 & 13.371 & 14.538 & 14.448 & 12.726 & 13.078 & 13.270 \\ \hline
SSP-LSTM \cite{chen2017long} & 8.951 & 8.227 & 10.757 & 8.938 & 11.503 & 12.896 & 6.526 & 8.763 & 10.767 \\ \hline
GR-LSTM & 6.086 & 8.636 & 10.801 & 9.295 & 11.116 & 12.978 & 6.484 & 8.707 & 10.614 \\ \hline
CSA-LSTM \cite{sun2019monaural} & 6.059 & 8.534 & 10.493 & 9.310 & 11.107 & 12.957 & 6.593  & 8.992 & 10.405\\ \hline
L2L \cite{ephrat2018looking} & 6.031 & 7.665 & 9.943 & 8.799 & 11.098 & 12.504 & 5.628 & 7.842 & 9.791 \\ \hline
OTF \cite{vaswani2017attention} & 5.820 & 7.889 & 10.197 & 8.747 & 11.038 & 13.005 & 5.768 & 8.319 & 10.247 \\ \hline\hline
\ms \ (ours) & \textbf{5.082} & \textbf{6.688} & \textbf{9.505} & \textbf{5.488} & \textbf{8.509} & \textbf{10.904} & \textbf{3.546} & \textbf{6.466} & \textbf{9.326} \\ \hline
\end{tabular}
\end{center}
\caption{\label{table:all}MSE bijection loss on the \dl \ test set for experiments in the Interclass, Intraclass and Hybrid partitions and $s \in \{2,3,5\}$. Lower number is better. For all $<u \in U, s \in S>$, \ms \ model shows superior performance in source separation than baselines.}
\end{table*} 

\subsection{Methodology}
In our experiments, models (\ms \ and baselines) unanimously take in a mixture ${X}_{<u, s, \cdot>}$, and output a predicated set $\bar{X}^{prd}_{<u, 1:s>}$ for $1:s$ sources (prd stands for prediction). The models are expected to predict the correct set of sources, but not necessarily in any particular order. This predicted set is subsequently compared against the ground-truth source set $\bar{X}_{<u, 1:s>}$. For comparison between these two sets, we use a greedy bijection with a similarity kernel as the comparison measure. This greedy approach simply maps each element in the predicted set to the most similar element in the ground-truth set, one after another. In our experiments, we use MSE as the similarity kernel. This constitutes fair comparison to baselines, since all the baselines originally train their models using MSE on the spectrogram predictions. The bijection approach used in this paper is summarized in Algorithm \ref{alg:model} in supplementary, and used for training \ms \ and baselines. After the loss is caluclated using this greedy bijection approach, it is normalized based on number of sources within the mixture for more meaningful comparison. 

Parameter optimization is done using Adam \cite{kingma2014adam} with learning rate~$\in \{0.001, 0.0005,0.0001\}$. All models are trained using dropout~$\in \{0,0.2,0.5\}$. Each model has its own specific tunable parameters as well. The number of LSTM layers in SSP-LSTM and GR-LSTM is $\in \{1,2,3\}$ layers with $\{32,64,128,256\}$ for hidden dimension shape. For STT and OTF, the number of encoders and decoders in both STT and OTF is $\{2,4,6,8\}$ layers and the number of heads in MSA is $\{1,2,4\}$. The hyperparameter space search of all the models was done using 12 Tesla V100 GPUs, for 1.5 months in duration.

\section{Results and Discussion}

\begin{table}[t!]
\fontsize{9}{10}\selectfont
\begin{center}
\renewcommand{\arraystretch}{1.2}
\begin{tabular}{K{4.00cm}|K{0.9cm}|K{0.9cm}|K{0.9cm}}
\hline
\multirow{2}{*}{\textit{Ablation Baseline}} & \multicolumn{3}{|c}{\textit{$u=$\textit{Hybrid}}} \\ \cline{2-4} 
 & $s=2$ & $s=3$ & $s=5$ \\ \hline
\ms \{tp-only\} & 6.020 & 8.582 & 10.103 \\ \hline
\ms \{sp-only\} & 6.557 & 8.502 & 10.311 \\ \hline
\ms \{tp-double\} & 5.696 & 8.398 & 10.194 \\ \hline
\ms \{sp-double\}& 6.115 & 8.394 & 10.363 \\ \hline
\ms \{no-CNN\}& 6.380 & 7.901 & 10.518 \\ \hline
\ms \{no-MGN\}& 5.516 & 7.189 & 10.229 \\ \hline
\ms & \textbf{3.563} & \textbf{6.683} & \textbf{9.578} \\ \hline
\end{tabular}
\end{center}
\caption{\label{table:ablation}Ablation experiments for STT model. The full STT model has the best performance compared to the ablation baselines. Therefore, all the components of the STT are necessary for achieving superior performance. }
\end{table}  
The results of our experiments are presented in Table \ref{table:all} for all the subdatasets $<u \in U, s \in S$. We summarize the observations from this table as following:\newline
 
\begin{figure}
    \centering
    \includegraphics[width=\linewidth]{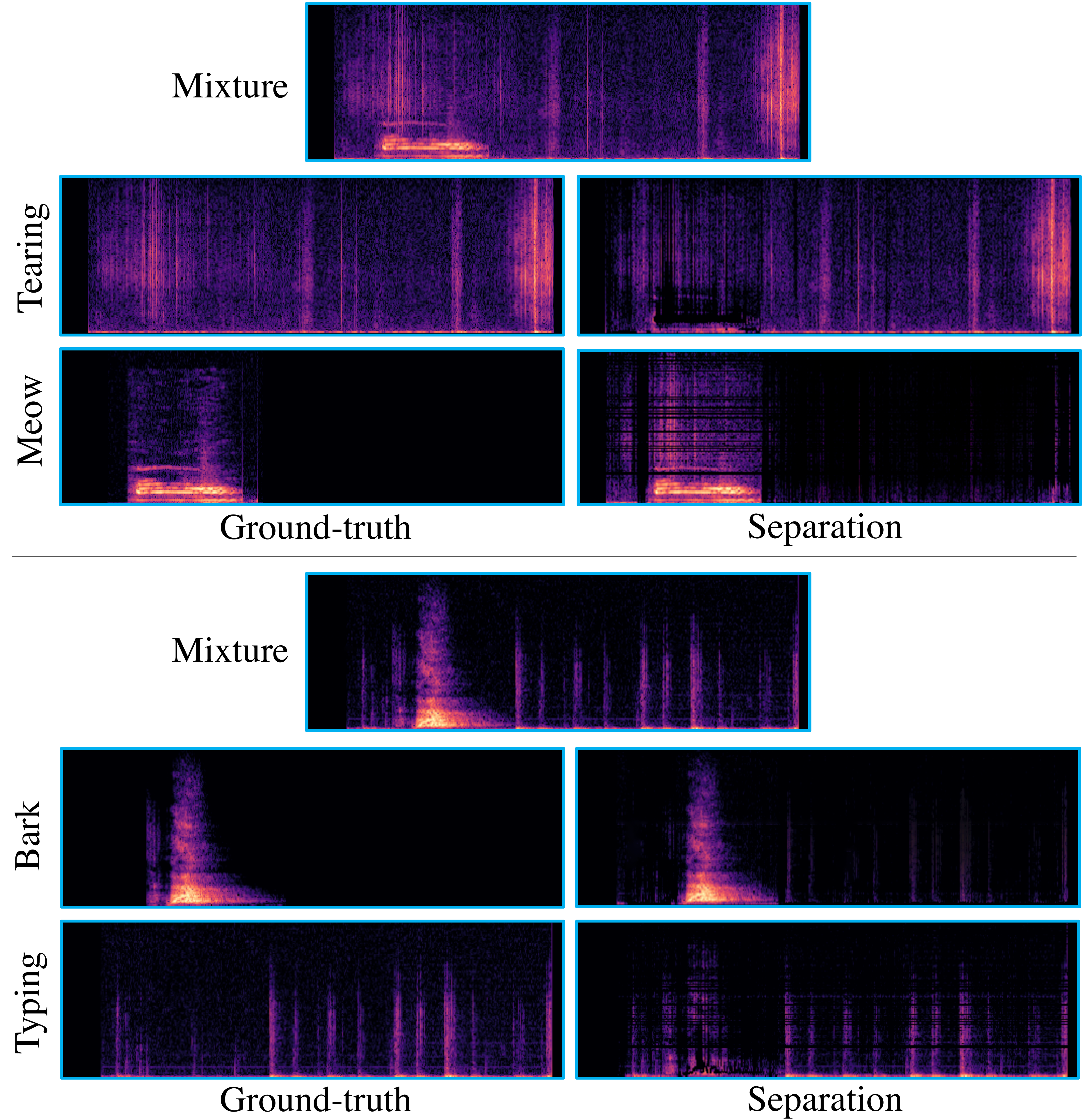}
    \caption{Example separation outputs of \ms \ for $s=2$ and $u=Hybrid$. STT is able to disentangle the sources within the given mixtures.}
    \label{fig:separations}
\end{figure}

\noindent \textbf{\ms \ Performance:}  In all the combinations of $U=\{$Interclass, Intraclass, Hybrid$\}$ and $S=\{2,3,5\}$, \ms \ achieves superior performance over the previously proposed models for source separation. RNN models (DRNN/SRNN) trail behind by a rather large margin. In contrast, LSTM-based models are able to achieve better performance than RNNs. Among LSTM-based approaches, L2L which uses combination of dilated convolutions and Bi-LSTM achieves the highest performance. OTF achieves superior performance than all LSTM-based models (except L2L). This demonstrates that the original transformer, even without specific designs for source separation, is more suitable for audio source separation than majority of RNN/LSTM models. Figure \ref{fig:separations} shows the qualitative performance of \ms, for $s=2$ and $u={Hybrid}$. Auditory separation examples are presented in supplementary. \newline

\noindent \textbf{Performance based on $S$:} Table \ref{table:all} shows that increasing the number of sources in the mixture naturally makes the problem of source separation more challenging. This is a consistent trend across all models in Interclass, Intraclass, and Hybrid partitions. \newline

\noindent \textbf{Performance based on $U$:} Table \ref{table:all} demonstrates that source separation in Intraclass partition is slightly more challenging than Interclass and Hybrid partitions. We believe this is due to the fact that sources across categories share less similarity, than sources within the same category. Therefore, naturally, it is harder to disentangle the mixtures in Intraclass partition. \newline

\noindent \textbf{\ms \ Ablation Studies:} To understand the importance of the tailored components of the STT model, we devise a set of ablation studies: 1) \textit{tp-only}, where we remove the spectral path and only keep the temporal path in STE. This is essentially the same as the original transformer encoder only with added convolutions. 2) \textit{sp-only}, where we keep only the spectral path and remove the temporal path in STE. 3) \textit{tp-double}, where the spectral path is replaced by a secondary temporal path in STE. 4) \textit{sp-double}, where temporal path is replaced by a secondary spectral path in STE. 5) \textit{no-CNN},  where the spectral and temporal path are present in STE but without CNNs. 6) \textit{no-MGN}, where the generation is a done using a simple feedforwards from decoder output, without masking. All these ablation baselines are compared for the Hybrid partition, which contains both Interclass and Intraclass elements. Table \ref{table:ablation} shows the results of this ablation experiment. The full \ms \ model achieves superior performance over the ablations. \newline

\section{Conclusion}

In this paper we presented a challenging new dataset for monoaural audio source separation, called \dl. \dl \ contains sounds from $25$ different classes, combined together using arbitrary start and volume, into mixtures. There are 9 subdatasets within the \dl, exactly combination of partitions $\{Interclass, Intraclass, Hybrid\}$ and tasks of $\{2,3,5\}$ source separation. We proposed a new transformer-based model for audio source separation called \ml \ (\ms). At the core of \ms, there is a specialized encoder called Spectro-Temporal Encoder (STE), which disentangles sources from across both temporal and spectral domains of the sound mixture. We compared the performance of the \ms \ to several previously proposed baselines for source separation over the \dl \ dataset. \ms \ showed superior performance in separating auditory sources across all the subdataset of \dl. As future direction, work has already started on \dl \ 2.0, which extends the number of classes to 100. 

\clearpage
\bibliography{aaai}
\bibliographystyle{aaai}

\clearpage
\begin{algorithm*}[b!]
\caption{\label{alg:mixture} Mixture creation algorithm for \dl \ dataset.}
\begin{algorithmic}[1]
\State $N_a \gets 60$, $N_T \gets 25$
\State ${A} \gets \{A_i;A_i=\{a_{<i,j>}\}_{i=1}^{N_a}\}_{j=1}^{N_T}$
\State $S \gets \{2,3,5\}$
\State $U \gets \{Interclass, Intraclass, Hybrid\}$
\State $F \gets \{tr,vl,te\}$ \quad $\triangleleft$ train, validation, test
\State $L \gets 2$ \quad $\triangleleft$ 2 Second Mixtures
\State $D_{s \in S, f=tr} \gets 10000 \times s$
\State $D_{s \in S, f \in \{vl,te}\} \gets 1000 \times s$
\Procedure{Mixture}{$A,s \in S,u \in U, f \in F, D_{s,f}, L$}
\State $X_{<u,s,f>} \gets []$ 
\For {$d=1 \dots D_{s,f}$}: 
\State $samples \in \textrm{SAMPLE-}u(A,s)$
\For {$m =1:s$}:
\State $st \sim uniform(0,L)$ \quad \quad $\triangleleft$  Random Start
\State $vol \sim uniform [\epsilon,1]$  \quad  $\triangleleft$ Random Volume
\State $\bar{X}_{<u,m,f>} \gets apply\{st, vol\}(samples[m])$ 
\EndFor
\State $X_{<u,s,f>} [i] \gets  \circledast \bar{X}_{<u,1:s,f>}$ \ $\triangleleft$  Mixing
\EndFor

\Return $X_{<u,s,f>}$
\EndProcedure
\Procedure{Sample-Intercategory}{$A, s \in S, f$}
\State $RC \xleftarrow[rand]{s} \{A\}_f$ \ \ $\triangleleft$ Random Sample $s$ Classes \\
\noindent \quad \ \ \Return $\cup_{i=1}^s \{rs \xleftarrow[rand]{1} RC[i]\}$
\EndProcedure

\Procedure{Sample-Intracategory}{$A, s \in S, f$}
\State $RC \xleftarrow[rand]{1} \{A\}_f$ \ \ $\triangleleft$ Random Sample $1$ Class \\
\noindent \quad \ \ \Return $rs \xleftarrow[rand]{s} RC$
\EndProcedure

\Procedure{Sample-Hybrid}{$A, s \in S, f$}\\
\noindent \quad \ \ \Return $rs \xleftarrow[rand]{s} \cup_{i=1}^s \{A\}_{f,i}$
\EndProcedure

\end{algorithmic}
\end{algorithm*}
\begin{algorithm*}[t!]
\caption{\label{alg:model} \ml \ (\ms) Formulation; for detailed implementation, please visit supplementary file code.zip. $tp$ stands for temporal and $sp$ stands for spectral. MSA stands for Multi-head Self Attention.} 
\begin{algorithmic}[1]
\State $\theta_{\ms} \gets \{\theta_{emb}\} \bigcup \{\theta_{STE}^{(i)}\}_{i=1}^{N_E} \bigcup \{\theta_D^{(i)}\}_{i=1}^{N_D} \bigcup \{\theta_{GR}\}$ \quad  \quad $\triangleleft$ Parameter Initialization
\State $X^{Em}_{<u, s,\cdot>}[i] \gets Embedding (X_{<u \ \in \ U, s \ \in \ S,\cdot>}[i])$ \quad  \quad $\triangleleft$ Embedding with positional information  \\
\Procedure{STE\_Stack}{$X^{E(0)}_{<\cdot>}[i],N_E$}
\For {$j = 0 \dots N_E-1$}:
\State $X^{E(j+1)}_{<\cdot>}[i] \gets STE\_Path(X^{E(j)}_{<\cdot>}[i],j,tp)+STE\_Path(X^{E(j)}_{<\cdot>}[i]^\intercal,j,sp)$
\EndFor\\
\quad \quad \Return $X^{E(N_E)}_{<\cdot>}[i]$
\EndProcedure
\Procedure{STE\_Path}{$X^{in,(j)}_{<\cdot>}[i],j,p \in \{tp, sp\}$}
\State $X^{CNN-p,(j)}_{<\cdot>}[i] \gets CNN^{p,(j)}(X^{in,(j)}_{<\cdot>}[i];\theta_{CNN^{p}}^{(j)})$
\State $X^{MSA-p,(j)}_{<\cdot>}[i] \gets MSA-p^{(j)}(X^{CNN-p,(j)}_{<\cdot>}[i];\theta_{MSA-p}^{(j)})$
\State $X^{MSA-p-AN,(j)}_{<\cdot>}[i] \gets Add\&Norm(X^{MSA-p,(j)}_{<\cdot>}[i], X^{CNN-p,(j)}_{<\cdot>}[i])$
\State $X^{FF-p,(j)}_{<\cdot>}[i] \gets FF-p^{(j)}(X^{MSA-p-AN,(j)}_{<\cdot>}[i])$
\State $X^{FF-p-AN,(j)}_{<\cdot>}[i] \gets Add\&Norm(X^{FF-p,(j)}_{<\cdot>}[i],X^{MSA-p-AN,(j)}_{<\cdot>}[i];\theta_{FF^p}^{(j)})$\\
\quad \ \ \Return $X^{FF-p-AN,(j)}_{<\cdot>}[i]$
\EndProcedure

\Procedure{Decoder\_Stack}{$X^{D(0)}_{<\cdot>}[i],N_D$}
\For {$j = 0 \dots N_D-1$}:
\State $X^{D(j+1)}_{<\cdot>}[i] \gets Decoder(X^{D(j)}_{<\cdot>}[i],j)$
\EndFor
\Return $X^{D(N_E)}_{<\cdot>}[i]$
\EndProcedure

\Procedure{Generation\_Residual}{$X^{D(N_D)}_{<\cdot>}[i]$}
\State $X^{FF1}_{<\cdot>}[i] \gets FF1(X^{D(N_D)}_{<\cdot>}[i])$
\State $X^{FF2}_{<\cdot>}[i] \gets FF2(X^{FF1}_{<\cdot>}[i])$\\
\quad \ \ \Return $X^{FF1}_{<\cdot>}[i] \times ReLU (X^{FF2}_{<\cdot>}[i])$
\EndProcedure

\Procedure{Greedy-Bijection-MSE}{$\bar{X}^{pred}_{<\cdot,1:s,\cdot>}, \bar{X}_{<\cdot,1:s,\cdot>}$}
\State $\forall (i,j)=1:s ; sim [i,j] \gets MSE (\bar{X}^{pred}_{<\cdot,i,\cdot>}, \bar{X}_{<\cdot,j,\cdot>})$
\State $\mathcal{L} \gets 0$
\For {$i = 1 \dots s$}:
\State $j \gets \argmin \big(sim[i,\cdot]\big) $
\State $\mathcal{L} = \mathcal{L} + sim [i,j]$
\State $\textrm{remove}\ sim [\cdot,j]$  \quad  \quad $\triangleleft$ Ensures Bijection
\EndFor
\Return $h_T, z_T$
\EndProcedure

\end{algorithmic}
\end{algorithm*}

\end{document}